\documentclass[10pt,twocolumn,letterpaper]{article}

\usepackage{soul}           % Strikethrough

\usepackage{cvpr}              % To produce the CAMERA-READY version
\usepackage[accsupp]{axessibility}

\definecolor{cvprblue}{rgb}{0.21,0.49,0.74}
\usepackage[pagebackref,breaklinks,colorlinks,allcolors=cvprblue]{hyperref}

\usepackage{multirow}

\def\paperID{14} % *** Enter the Paper ID here
\def\confName{CVPR}
\def\confYear{2025}

\title{Enhancing Vision Transformer Explainability Using Artificial Astrocytes}

\author{Nicolas Echevarrieta-Catalan\\
University of Miami\\
Coral Gables, FL, USA\\
{\tt\small nxe272@miami.edu}
\and
Ana Ribas-Rodriguez\\
University of A Coruña\\
A Coruña, Spain\\
{\tt\small ana.ribas@udc.es}
\and
Francisco Cedron\\
University of A Coruña\\
A Coruña, Spain\\
{\tt\small francisco.cedron@udc.es}
\and
Odelia Schwartz\\
University of Miami\\
Coral Gables, FL, USA\\
{\tt\small odelia@cs.miami.edu}
\and
Vanessa Aguiar-Pulido\\
University of Miami\\
Coral Gables, FL, USA\\
{\tt\small vanessa.aguiar@miami.edu}
}

\begin{document}
\maketitle
\begin{abstract}
% Just some text for reservation of space: 

% It began with the forging of the Great Rings. Three were given to the Elves, immortal, wisest and fairest of all beings. Seven to the Dwarf-Lords, great miners and craftsmen of the mountain halls. And nine, nine rings were gifted to the race of Men, who above all else desire power. For within these rings was bound the strength and the will to govern each race. But they were all of them deceived, for another ring was made. Deep in the land of Mordor, in the Fires of Mount Doom, the Dark Lord Sauron forged a master ring in secret, and into this ring he poured his cruelty, his malice and his will to dominate all life. One ring to rule them all.

    Machine learning models achieve high precision, but their decision-making processes often lack explainability. Furthermore, as model complexity increases, explainability typically decreases. Existing efforts to improve explainability primarily involve developing new eXplainable artificial intelligence (XAI) techniques or incorporating explainability constraints during training. While these approaches yield specific improvements, their applicability remains limited. In this work, we propose the Vision Transformer with artificial Astrocytes (ViTA). This training-free approach is inspired by neuroscience and enhances the reasoning of a pretrained deep neural network to generate more human-aligned explanations. We evaluated our approach employing two well-known XAI techniques, Grad-CAM and Grad-CAM++, and compared it to a standard Vision Transformer (ViT). Using the ClickMe dataset, we quantified the similarity between the heatmaps produced by the XAI techniques and a (human-aligned) ground truth. Our results consistently demonstrate that incorporating artificial astrocytes enhances the alignment of model explanations with human perception, leading to statistically significant improvements across all XAI techniques and metrics utilized.
\end{abstract}    
\section{Introduction}
\label{sec:introduction}    

    Machine learning has demonstrated its ability to perform as well as or better than humans in a wide range of tasks by learning from data. However, machine learning models are frequently seen as a "black box". Understanding how the model reasons is of paramount importance in domains of application like the medical domain. One of the main goals of eXplainable Artificial Intelligence (XAI) is to obtain human-understandable explanations or interpretations that shed light on how a machine learning model reasons. In the field of computer vision, explanations often take the form of a heatmap, highlighting which parts of an image the model considers most important. Ideally, the pixels highlighted by the model should coincide with those that a human considers important for classifying an image into a certain category. One of the most widely used families of methods for XAI in image classification is Class Activation Mapping (CAM) \cite{zhou2016learning}. Although CAM-based methods were originally developed for Convolutional Neural Networks (CNNs), they have since been adapted for use with the Transformer architecture \cite{fantozzi2024explainability}. CAM-based methods generate a heatmap by projecting the predicted class scores onto the input image, visually highlighting the importance of specific pixels in the image that are most relevant to the model's prediction for the selected class.

    As computer vision models become more complex, their explanations often become less interpretable to humans. This occurs because highly optimized models rely on patterns that may not align with human reasoning, making their decision-making difficult to validate. Ideally, relevant information in an image that determines an object's presence should be independent of the observer, whether human or artificial intelligence (AI). In a human-aligned system, both humans and AI should rely on similar visual cues \cite{samek2017explainable}. Addressing this challenge requires a shift towards human alignment, which refers to the degree to which the explanations of an AI model align with human reasoning \cite{doshi2017towards}. Ensuring human alignment is essential in fields where interpretability and trust are critical, such as healthcare, autonomous systems, and legal decision-making.

    In this work, we drew inspiration from neuroscience to develop a novel deep learning approach that enhances the reasoning of a pretrained neural network to generate more human-aligned explanations, enhancing already existing explainability methods. We hypothesize that integrating modulation processes inspired by those in the human brain into vision-based deep neural networks could enhance explainability. Specifically, we introduced astrocytes—a type of glial cell involved in synaptic processes—into the first self-attention block of a Vision Transformer, and explored their impact on explainability methods in computer vision. We focused on astrocytes because of their ability to enhance and inhibit neural activity. We therefore asked whether this modulation process could highlight visual information in a way that is more aligned with humans.

    We compared the explanations generated employing the original Vision Transformer (ViT) and the proposed approach—Vision Transformer with artificial Astrocytes (ViTA), against human relevance ground truth from the ClickMe dataset \cite{linsley2019learning}. Our results indicate that explanations obtained using ViTA are significantly more aligned with human relevance than those generated utilizing ViT. The novelty of this approach lies in the biologically inspired and method-agnostic enhancement of explainability through the inclusion of astrocytes. To the best of our knowledge, this has not been explored before.
\section{Related work}
\label{sec:related_work}

    Astrocytes have been utilized in artificial intelligence for some time, from their initial incorporation into multilayer perceptrons (MLPs) \cite{ikuta2009multi, porto2011artificial}, to more recent applications in CNNs \cite{ribas2024training}, spiking neural networks \cite{gordleeva2021modeling}, and dual neuron-astrocyte networks \cite{han2023astronet, han2024ma}. Additionally, Kozachkov \etal \cite{kozachkov2023building} modeled the internal mechanisms and outputs of a Transformer block using astrocytes, observing that the self-attention could be replaced by a neuron-astrocyte model. Notably, all these applications have primarily focused on classification tasks, leaving their potential to enhance explainability largely unexplored.

    To enhance the explanations of ViTs, researchers have combined gradient-based and attention-based explanation methods. For example, Chefer \etal \cite{chefer2021transformer} integrated relevance and attention scores to generate explanations, while Brocki \etal \cite{brocki2023class} used the gradients from the prediction to scale the attention scores based on the relevance of the corresponding token in the model’s decision. Alternatively, other approaches aim to improve the explanations by incorporating them into the training process. Fel \etal \cite{fel2022harmonizing} trained a ViT with an additional explainability term in the loss function, while Kang \etal \cite{kang2025improving} introduced a parallel Patch-level Mask prediction module, jointly trained with the classifier head. It is worth noting that these explainability approaches are often tied to specific ViT implementations and do not enhance pre-existing methods, limiting their usability. In this paper, we propose a novel, training-free approach to improve the output of XAI methods by incorporating artificial astrocytes into the multi-head self-attention mechanism of a ViT.
\section{Materials and methods}
\label{sec:materials_and_methods}

% \subsection{Vision Transformer}

%     The Vision Transformer (ViT) \cite{dosovitskiy2020image} is an adaptation of the Transformer architecture for image classification. The input image is converted into tokens by splitting it into patches, flattening those patches, and adding a positional embedding. Additionally, an extra class token (CLS) is added at position 0. This CLS token aggregates the information required for classification and serves as the input to the classifier head after all the attention blocks.

%     Each attention block consists of a multi-head self-attention module followed by a two-layer Multi-Layer Perceptron (MLP). Both modules are preceded by a normalization layer and connected to the residual stream via a residual connection, as shown in \cref{fig:ViT_architecture}. The multi-head self-attention module computes the relationships between the Query, Key, and Value representations for all tokens and applies a linear transformation to the output before merging it with the residual stream.

In this work, we devised a neuroscience-inspired Vision Transformer with artificial Astrocytes (ViTA). Artificial astrocytes are incorporated into a pretrained ViT, thereby eliminating the need for network training; instead, only the optimization of astrocytic hyperparameters is required. The astrocytic modulation of neurons allows enhancing the output of XAI techniques.

\subsection{Vision Transformer with artificial Astrocytes (ViTA)}

\begin{figure}[!ht]
        \centering
        \includegraphics[width=0.8\linewidth]{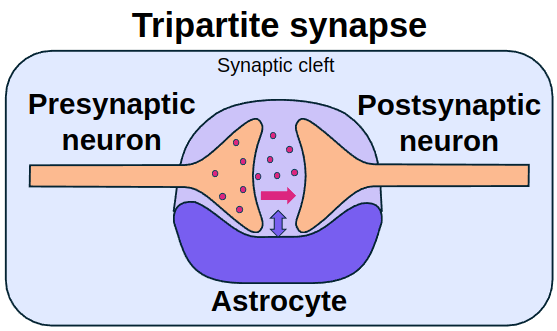}
    
        \caption{Illustration of the tripartite synapse. Information is transmitted along the neural pathways as the presynaptic neuron sends neurotransmitters to the postsynaptic neuron. The astrocyte surrounds the synaptic cleft (i.e., the space between neurons). Astrocytes interact and help to modulate this process in the so-called tripartite synapse.}
        \label{fig:tripartite_synapses}
    \end{figure}

    Biological astrocytes \cite{ARAQUE1999208} regulate neuronal synapses in response to synaptic neuron activity \cite{saint2023astrocyte}, enabling synaptic plasticity. This mechanism modulates neurotransmitter levels in the synaptic cleft, thereby influencing the signal transmitted to the postsynaptic neuron and forming tripartite synapses, as illustrated in \cref{fig:tripartite_synapses}. Astrocytic influence on synapses can be either excitatory or inhibitory \cite{vasile2017human}, and this process operates on significantly slower timescales than neuronal transmission \cite{vardjan2016loose}. Here, we designed artificial astrocytes based on three key aspects of biological astrocytes: excitatory modulation, inhibitory modulation, and differing timescales. %We incorporated these artificial astrocytes into a Vision Transformer pretrained on ImageNet and evaluated the impact on CAM-based explainability methods. 
    % Biological astrocytes regulate neuronal synapses in response to synaptic neuron activity \cite{saint2023astrocyte}, enabling synaptic plasticity. This mechanism modulates neurotransmitter levels in the synaptic cleft, thereby influencing the signal transmitted to the postsynaptic neuron and forming tripartite synapses. Specifically, astrocytes can release CA2+ ions, stimulating the presynaptic neuron to release more neurotransmitters, leading to an excitatory effect. Conversely, they can also reabsorb neurotransmitters from the synaptic space, producing an inhibitory effect \cite{vasile2017human}

\begin{figure}[t]
        \centering
        \includegraphics[width=0.8\linewidth]{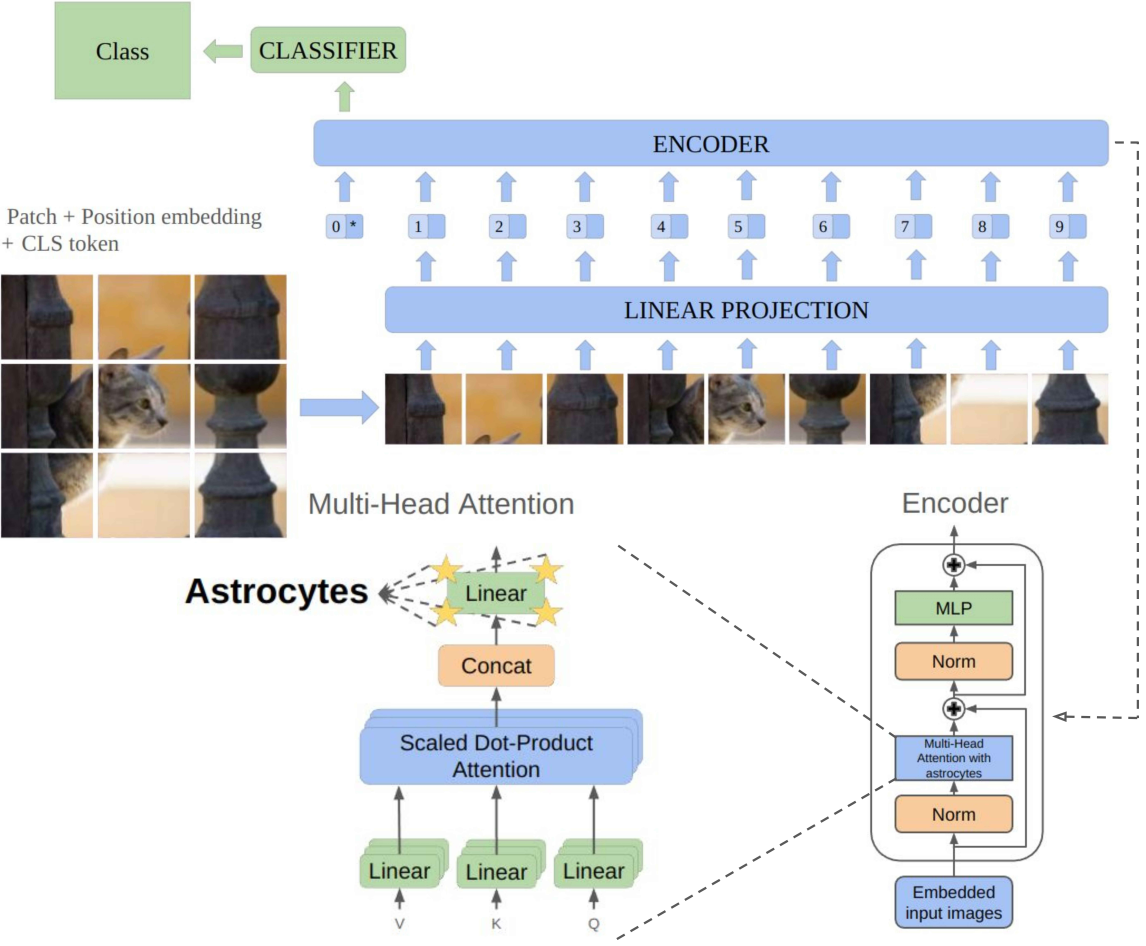}
    
        \caption{Proposed architecture: Vision Transformer with artificial Astrocytes (ViTA). Artificial astrocytes (stars) are added to the dense layer of the multi-head attention module of the encoder block 0.}
        \label{fig:ViT_architecture}
    \end{figure}

    The proposed architecture adds astrocytes to the first attention block of the ViT (see \cref{fig:ViT_architecture}). Similarly to biological astrocytes, artificial astrocytes modulate the activity of artificial neurons and do so within the linear layer of the first attention block. Each artificial astrocyte further excites or inhibits the signal being transmitted by the presynaptic artificial neuron depending on its activation level over multiple iterations. Given the timescale differences between neurons and astrocytes \cite{vardjan2016loose}, we implemented the astrocytic modulation as an iterative process. This iterative process applies only to the linear layer at the end of the first self-attention block, minimizing additional computations required for the astrocytic modulation. Each image passes through the network only once; however, when the processing reaches the layer that includes artificial astrocytes, the input to that layer iterates \(k\) times, increasing the effect of the modulation. After the last iteration, a single (modulated) output is generated and passed to the next processing unit of the model, which continues operating as usual. It is worth highlighting that placing the artificial astrocytes in the first decoder block will maximize their influence throughout the network. %As the astrocytes are in the first decoder block, their modulation will be applied to an input with less processing from the original image, transmitted to the rest of the network, maximizing its effect over the network.
    We used a 1:1 ratio of neurons to astrocytes based on previous work \cite{kozachkov2023building} and, for simplicity, we did not include inter-astrocyte communication in our approach. As a result, the modulation of each neuron is independent of the modulation of the other neurons within the same layer. Note that the astrocytic linear layer should only be used for explainability and not classification. Additional details are provided subsequently.

    \subsubsection{Astrocytic parameters}

        The behavior of the artificial astrocytes in ViTA varies depending on five parameters:

        \begin{itemize}
            \item \textbf{Number of iterations} (\(k\)) represents the different time scales on which neurons and astrocytes operate.  
            \item \textbf{Response speed} (\(\tau\)) determines how quickly the astrocyte responds to neuronal activity.  
            \item \textbf{Activation level threshold} (\(\phi\)) defines the astrocyte's sensitivity to the presynaptic neuron's activation level.  
            \item \textbf{Excitatory modulation factor} (\(\alpha\)) regulates the intensity of the excitatory modulation.  
            \item \textbf{Inhibitory modulation factor} (\(\beta\)) controls the intensity of the inhibitory modulation.  
        \end{itemize}

    \subsubsection{Astrocytic linear layer implementation}

        As previously stated, ViTA incorporates artificial astrocytes into the first self-attention block, replacing its linear layer with an astrocytic linear layer. In the ViT architecture, the linear layer within the self-attention block applies the same weights and bias to all tokens that are generated from the input image. We chose to use only the layer's output CLS token to determine each neuron's activation level, as it aggregates information from all other tokens and is ultimately transmitted to the classifier head after the final self-attention block. Rather than modulating the weights for each token individually, the modulation induced by the CLS token affects all tokens collectively.

        For a linear layer without astrocytes, the activation level of a neuron \( i \) is computed by multiplying the layer's input \( x \) by a set of weights \( W \), followed by the addition of a bias \( b \):  

        \begin{equation}
            y_{i} = xW_{i}^{T} + b_{i}
        \end{equation}
        
        The behavior of this linear layer was modified to include an astrocytic modulation through an iterative process. Unlike for the standard ViT model, in the proposed model each input (i.e., image) is presented \(k\) times to the astrocytic linear layer, which represents the longer timescale required for astrocyte-neuron communication. Hence, in each iteration, the input image is processed similarly to the standard ViT model until it reaches the astrocytic linear layer. In this layer, the weights \( W \) are multiplied by a modulation factor \( M \), which accumulates over time (i.e., over a total of \(k\) iterations). As a result, the modulated activation level (output) of neuron \(i\) at iteration \(t\) is given by:  
        
        \begin{equation}
            y_{i}(t) = x(M(t) W)_{i}^{T} + b_{i}
        \end{equation}

        The modulation factor \( M \) is a diagonal matrix initialized as the identity matrix, where the position $ii$ is the accumulated modulation for neuron $i$, and increases at each iteration by a factor \( m_{i} \):

        \begin{equation}
            M(0) = I
        \end{equation}  
        \begin{equation}
            M_{ii}(t) = M_{ii}(t-1) \cdot m_{i}(t)
        \end{equation}

        where \( m_{i} \) can take the values \( \alpha \geq 1 \) for an excitatory modulation, \( 0 < \beta < 1 \) for an inhibitory modulation, or \( 1 \) when no modulation condition is met. These modulation conditions are defined by the presynaptic neuron's activation level over previous iterations. If the neuron has been primarily active (i.e., active for at least $\tau$ iterations), the astrocyte induces an excitatory modulation. On the other hand, if the neuron has been primarily "inactive" (i.e., "inactive" for at least \(\tau\) iterations, which we denote as \(-\tau\)), the astrocyte would generate an inhibitory modulation. When neither condition is met, no variation in the modulation occurs. To determine whether a modulation condition is satisfied, the activity of the neuron $i$ over the iterations is tracked by $A_{i}$.

        \begin{equation}
            m_{i}(t) =
            \begin{cases}
                \alpha, & \text{if } A_{i}(t) \ge \tau \\
                \beta, & \text{if } A_{i}(t) \leq -\tau \\
                1, & \text{otherwise}
            \end{cases}
        \end{equation}
        
         $A_{i}$ is initialized to 0, is bounded by the interval \([- \tau, \tau]\), and is updated based on the activation level of the presynaptic neuron \(y_{i}\) during previous iterations. If the activation level of the neuron is greater than or equal to an activation level threshold \( \phi \), then $A_{i}$ will increase by +1. Conversely, if it is below \( \phi \), $A_{i}$ decreases by -1. 

        \begin{equation}
            A_{i}(0) = 0
        \end{equation}  
        \begin{equation}
            A_{i}(t) =
            \begin{cases}
                \tau, & \text{if } a_{i}(t) \ge \tau \\
                -\tau, & \text{if } a_{i}(t) \leq -\tau \\
                a_{i}(t) & \text{otherwise}
            \end{cases}
        \end{equation}
        % \begin{equation}
        %     a_{i}(t) = A_{i}(t-1) + \operatorname{sign}(y_{i}(t-1) - \phi)
        % \end{equation}         

        being \(a_{i}\) the updated value of \(A_{i}\) before constraining to the interval \([-\tau, \tau]\):
        
        \begin{equation}
            a_{i}(t) =
            \begin{cases}
                A_{i}(t-1) + 1, & \text{if } y_{i}(t-1) \ge \phi \\
                A_{i}(t-1) - 1, & \text{otherwise}
            \end{cases}
        \end{equation}

        After the iterative process has concluded, the output from the astrocytic linear layer is normalized to match the scale of a standard linear layer's output. Thus, the final output \( \hat{y}(k) \) is obtained by multiplying the output of the last iteration \( y(k) \) by the ratio of the mean norms of the output for each token, computed without modulation (\( t=0 \)) and at the last iteration (\( t=k \)):  

        \begin{equation}
            \hat{y}(k) = y(k) \cdot \frac{\operatorname{mean} \left( \left[ \left\| y_i(0) \right\|_{2} \right] \right)}
            {\operatorname{mean} \left( \left[ \left\| y_i(k) \right\|_{2} \right] \right)}, \quad \forall i \text{ in tokens}
        \end{equation}

        After normalization, the output is processed in the same manner as it would be in a standard ViT, propagating the effect of the astrocytic modulation through the rest of the network and through the residual stream.

\subsection{Explainability}
    
    Different explainability methods can be applied to Transformers to obtain an explanation of its reasoning \cite{fantozzi2024explainability}. In this work we focus on a widely used family of XAI techniques: the Class Activation Mapping (CAM) family \cite{zhou2016learning}. We employed two well-established methods from the pytorch\_grad\_cam library \cite{jacobgilpytorchcam}: \textit{Grad-CAM} \cite{selvaraju2016grad} and \textit{Grad-CAM++} \cite{chattopadhay2018grad}. The key difference between the two techniques is that Grad-CAM's activation mappings are weighted by the average gradient during the backward pass, whereas Grad-CAM++ uses second-order gradients for weighting.

\subsection{Dataset - ClickMe}

    We utilized human heatmaps from the ClickMe dataset \cite{linsley2019learning} as ground truth to assess our model's explainability. This dataset is a subset of ImageNet ILSVRC12 that includes human relevance heatmaps, which we used to evaluate the alignment of explanations provided by XAI techniques using ViT and ViTA with human perception. Because it is a subset of ImageNet, no fine tuning or training is needed. Thus, we used ViT's public weights pretrained on ImageNet from the \textit{timm} deep learning library \cite{wightman2019timm}.

    % ClickMe relevance heatmaps were generated through a game in which images from ImageNet were presented to human players. Each player clicked on the parts of the image they considered most relevant for human recognition, effectively 'painting' them. As the image was painted, the highlighted regions were copied onto a white canvas, which was then shown to a deep convolutional network (DCN) pretrained on ImageNet. The DCN attempted to predict the class of the image, with players earning more points the faster the DCN correctly recognized the class.
    
    % ClickMe was preceded by another game, Clicktionary \cite{linsley2017visual}, in which the adversarial player was another human instead of a DCN. However, this game was discontinued after collecting heatmaps for a few hundred images in favor of the Human-DCN approach of ClickMe, as the authors found it impractical to scale the matching of player pairs effectively. Nevertheless, the heatmaps obtained from Clicktionary were used to validate ClickMe by comparing results for the same images, revealing a strong correlation between them. Additionally, ClickMe heatmaps were further validated against players using DCN-specific strategies to obtain high scores.
    
    % The areas 'painted' by 1,235 players (identified by unique user IDs) across 25 competitions were analyzed, producing heatmaps for each image. The intensity of a pixel in these maps represents the probability that it was selected (or 'painted') by a player as relevant for recognition.
    
\begin{figure}[t]
        \centering
        \includegraphics[width=0.8\linewidth]{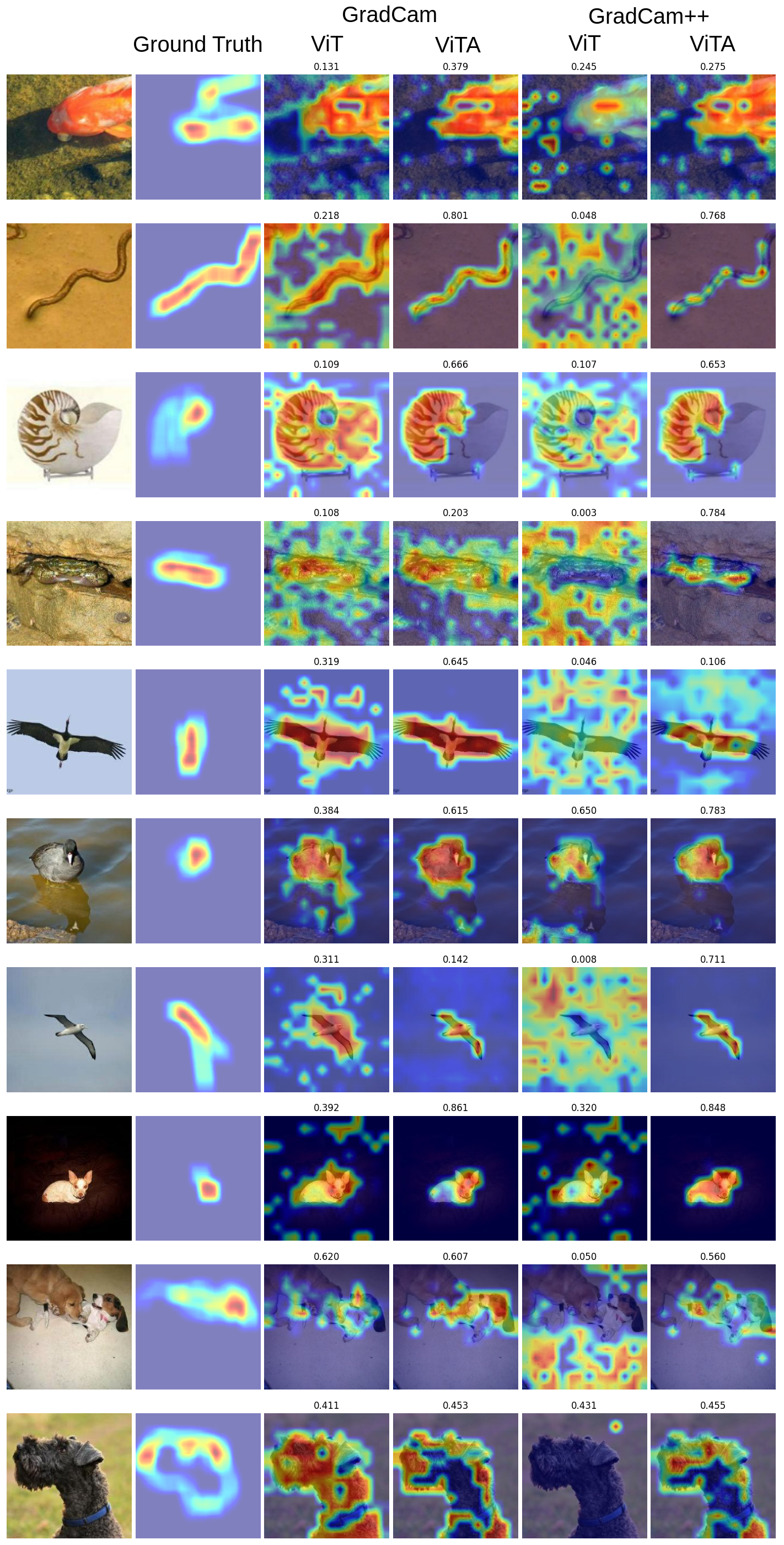}
    
        \caption{Class activation maps produced by Grad-CAM and Grad-CAM++ for ViT and ViTA. The columns correspond to the: (1) original image, (2) (human-aligned) ground truth, Grad-CAM output for (3) ViT and (4) ViTA, and Grad-CAM++ output for (5) ViT and (6) ViTA. The numerical values above the images in columns 3–6 represent SSIM scores, indicating how closely the heatmaps generated by each method align with the ground truth.}
        \label{fig:explainability_examples}
    \end{figure}

    Due to the dataset's imbalance and the presence of duplicated images with different heatmaps, we randomly selected 2,982 unique images from the ClickMe validation set: three images for each of the 1,000 ImageNet classes, except for two classes, which contained only one image, and fourteen classes that contained only two unique images. Examples of these images and their (human) ground truth heatmaps are shown in \cref{fig:explainability_examples}.

\subsection{Evaluation}

    To measure the overlap between the activation maps obtained using the two XAI techniques and the human ground truth, we employed the following well-established metrics \cite{fel2022harmonizing, malladi2023towards}:

    \begin{description}

        % \item[Trace (or Frobenius) inner product]

        \item[Spearman correlation \cite{spearman1904proof}] is a nonparametric measure of rank correlation. It assesses how well the relationship between two variables can be described using a monotonic function. The result lies in the interval [-1, 1], where +1 indicates identical ranks and -1 indicates inverse ranks.
        
        \begin{equation}
            Spearman (x, y) = \frac{COV(R[x], R[y])}{\sigma_{R[x]}\sigma_{R[y]}}
        \end{equation}

        % \item[Intersection over union (or Jaccard index)] measures similarity between finite non-empty sample sets. A perfect fit will have a IOU of 1, and the most different both sets are, the closer to 0 IUO will be.

        % \begin{equation}
        %     IOU (x, y) =  \frac{|x \cap y|}{|x \cup y|}
        % \end{equation}

        \item[Dice Similarity Coefficient (DSC) \cite{dice1945measures}] measures the similarity between finite, non-empty sample sets. A perfect match results in a DSC of 1, while greater dissimilarity between the sets brings the DSC closer to 0.

        \begin{equation}
            DSC (x, y) =  \frac{2*|x \cap y|}{|x| + |y|}
        \end{equation}

        \item[Structural similarity (SSIM) \cite{wang2004image}] was designed to assess quality degradation in digital images. It evaluates structural information within the images by combining statistics of the pixels in different directions. SSIM ranges from $[-1, 1]$, where 1 indicates a perfect match (identical images), -1 represents completely opposite images, and 0 signifies entirely unrelated images. 

        \begin{equation}
            SSIM(x, y) = \frac{(2\mu_x\mu_y + C_1)(2\sigma_{xy} + C_2)}{(\mu_x^2+\mu_y^2+C_1)(\sigma_x^2+\sigma_y^2+C_2)}
        \end{equation}

        where $C_1$ and $C_2$ are two constants to stabilize the division when the denominator is weak.
        
        % \iten[ Percentage of Correct Keypoint (PCK)]

        % \item[Hausdorff distance?]
        
    \end{description}

\section{Results}

    First, we conducted a grid search to identify the optimal astrocytic configuration. Parameter values that maximized overlap of the heatmaps (activation maps) generated employing the different XAI techniques with the ground truth were chosen. The values considered for each parameter during the grid search are included below:
    \begin{itemize}
        \item \textbf{Number of iterations} (\(k\)): [4, 6, 8]
        
        \item \textbf{Activation number} (\(\tau\)): [1, 2, 3]
        
        \item \textbf{Activation level threshold} (\(\phi\)): [-0.5, -0.2, 0.0, 0.2, 0.5]
        
        \item \textbf{Excitatory modulation factor} (\(\alpha\)): [1.05, 1.2, 1.5]
        
        \item \textbf{Inhibitory modulation factor} (\(\beta\)): [0.005, 0.05, 0.25]
    \end{itemize}

    The best configuration for each XAI technique and metric is provided in \cref{tab:ViTA_metric_comparison_part1}.

    \begin{table*}[!htbp]
        \centering
        \begin{tabular}{@{}|c|c|c|c|c|c|c|@{}} \hline
            \multirow{2}{*}{CAM} & \multirow{2}{*}{Metric} & \multirow{2}{*}{k} & \multirow{2}{*}{$\tau$} & \multirow{2}{*}{$\phi$} & \multirow{2}{*}{$\alpha$} & \multirow{2}{*}{$\beta$} \\ 
            &  &  &  &  &  & \\ \hline
            Grad-CAM & Spearman & 8 & 1 & 0.2 & 1.25 & 0.005 \\ \hline
            Grad-CAM & DSC & 4 & 3 & -0.5 & 1.25 & 0.05 \\ \hline
            Grad-CAM & SSIM & 6 & 3 & -0.5 & 1.5 & 0.05 \\ \hline
            Grad-CAM++ & Spearman & 6 & 3 & -0.5 & 1.25 & 0.25 \\ \hline
            Grad-CAM++ & DSC & 4 & 3 & -0.5 & 1.5 & 0.005 \\ \hline
            Grad-CAM++ & SSIM & 8 & 1 & -0.5 & 1.5 & 0.005 \\ \hline
        \end{tabular}
        \caption{Best parameter configurations for ViTA}
        \label{tab:ViTA_metric_comparison_part1}
    \end{table*}
    
\begin{figure}[t]
        \centering
        \includegraphics[width=0.95\linewidth]{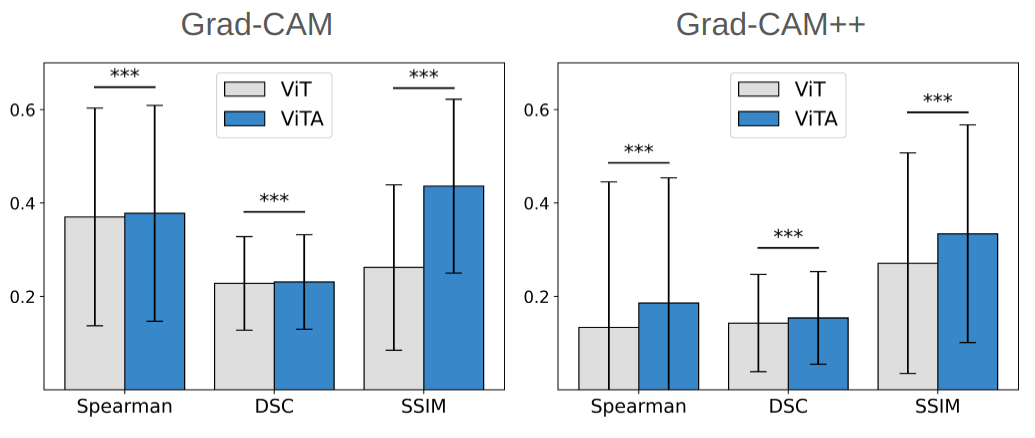}

        \caption{Comparison of Grad-CAM and Grad-CAM++ applied to ViT and ViTA against ClickMe ground truth using Spearman, DSC and SSIM with the best parameter configurations from \cref{tab:ViTA_metric_comparison_part1}. Error bars represent variability.}
        
        \label{fig:barplot_cam_metric_comparison}
    \end{figure}

    Next, Grad-CAM and Grad-CAM++ were utilized to generate visual explanations in the form of heatmaps for both ViT and ViTA. The images obtained were compared employing three different metrics: Spearman correlation, DSC and SSIM. Finally, a one-tailed Wilcoxon rank-sum test was employed to evaluate statistical significance. The null hypothesis stated that both methods performed equally, while the alternative hypothesis posited that ViTA achieved greater alignment with the human ground truth. \cref{tab:ViTA_metric_comparison_part2} shows the results of evaluating the similarity between the human-aligned heatmaps and those generated by each CAM-based XAI technique and transformer architecture used. The mean, median and standard deviation (SD), along with statistical significance values are included. A visual representation of the difference in explainability is illustrated in \cref{fig:barplot_cam_metric_comparison}. \cref{fig:explainability_examples} includes examples of the astrocytic modulation effect on activation maps.

        \begin{table*}[!htbp]
        \centering
        \begin{tabular}{@{}|c|c|c|c|c|c|c|c|c|@{}} \hline
            \multirow{2}{*}{CAM} & \multirow{2}{*}{Metric} & \multicolumn{3}{c|}{ViT} & \multicolumn{3}{c|}{ViTA} & \multirow{2}{*}{p-value} \\ 
            &  & Mean & Median & SD & Mean & Median & SD &  \\ \hline
            Grad-CAM & Spearman & 0.370  & 0.401  & 0.233 & \textbf{0.378} & 0.401 & 0.231 & \textbf{1.9e-08}*** \\ \hline
            Grad-CAM & DSC & 0.228  & 0.220  & 0.100 & \textbf{0.231} & 0.224 & 0.101 & \textbf{1.9e-14}*** \\ \hline
            Grad-CAM & SSIM & 0.262  & 0.225  & 0.177 & \textbf{0.436} & 0.432 & 0.186 & \textbf{3.9e-290}*** \\ \hline
            Grad-CAM++ & Spearman & 0.134  & 0.148  & 0.311 & \textbf{0.186} & 0.200 & 0.268 & \textbf{9.2e-13}*** \\ \hline
            Grad-CAM++ & DSC & 0.143  & 0.128  & 0.104 & \textbf{0.154} & 0.142 & 0.099 & \textbf{6.4e-09}*** \\ \hline
            Grad-CAM++ & SSIM & 0.271  & 0.183  & 0.236 & \textbf{0.334} & 0.303 & 0.233 & \textbf{4.2e-28}*** \\ \hline
        \end{tabular}
        \caption{Similarity of heatmaps produced by Grad-CAM and Grad-CAM++ for ViT and ViTA with the ClickMe (human-aligned) ground truth using Spearman, DSC and SSIM. Mean, median, standard deviation (SD), and p-value for the difference of the means.}
        \label{tab:ViTA_metric_comparison_part2}
    \end{table*}

    Results show that the best ViTA configurations significantly improve explainability, regardless of the XAI technique and metric used to assess similarity with human-aligned ground truth. Notably, SSIM is the metric that shows the highest improvement when using Grad-CAM. The best parameter configurations for the astrocytes are those with a strong excitatory ($\alpha$ of $1.25$ or $1.5$) and inhibitory ($\beta$ of $0.05$ or $0.005$) modulations, paired with a low activation level threshold ($\phi$ of $-0.5$), except for Grad-CAM and Spearman ($\phi$ of $0.2$). These configurations correspond to a highly sensitive astrocyte with excitatory tendency. On the other hand, the optimal combinations for the number of iterations ($k$) and response speed ($\tau$), which represent the time scale difference and reaction speed respectively, exhibit greater variability. %Upon examination of the activation maps (\cref{fig:explainability_examples}), ViTA appears to enhance these heatmaps by focusing more on the object in the image while minimizing background noise, resulting in a more accurate alignment with the object of interest, and consequently, the human relevance ground truth.
   
    Upon examination of the activation maps (\cref{fig:explainability_examples}), ViTA's astrocytes appear to be accentuating image content that leads to stronger activations while suppressing content that leads to weaker activations. By reducing noise and emphasizing relevant content, the information entering the residual stream becomes more focused on stronger activations. These activations are then propagated through the model up to the last attention block, where the explanation is extracted. This results in a more accurate alignment with the object of interest, and consequently, the human ground truth.

\section{Conclusions and Future Work}
\label{sec:conclusions}

   In this work, we propose a novel approach that employs a modulation mechanism inspired by biological astrocytes to achieve better explainability. Our results, evaluated across multiple metrics, demonstrate that incorporating artificial astrocytes into the first self-attention block improves the alignment of model explanations with human ground truth when using gradient-based CAM methods. Specifically, the explanations obtained show greater overlap with human relevance maps. Moreover, the heatmaps generated by ViTA appear to be more focused on the object in the image, while minimizing attention to background regions, further demonstrating the effectiveness of astrocytic modulation in enhancing explainability.

    % In this work, we created a novel approach that, by using a modulation mechanism inspired from how biological astrocytes modulate neuronal synapses, allows to obtain more human-aligned explanations from a pretrained Vision Transformer model. The results from different metrics support that, by adding astrocytes to the self-attention block of a vision transformer, the explanations obtained using different XAI approaches overlap better the human ground truth when using gradient-based CAM methods than without the astrocytes. When looking at the explanations themselves, the obtained heatmaps seem more focused on the object in the image and generally are less focused on the background.
    
    Future work will involve incorporating astrocytes into other transformer architectures (e.g., DINOv2), exploring additional XAI techniques, and utilizing a variety of segmentation datasets across different application domains. Finally, we will employ additional metrics to evaluate the proposed approach's performance on various types of images.

{
    \small
    \bibliographystyle{unsrt}
     \bibliography{main}
    %\bibliography{bibliography}
}


\end{document}